# Spatio-Temporal Denoising Graph Autoencoders with Data Augmentation for Photovoltaic Timeseries Data Imputation

Yangxin Fan, Xuanji Yu, Raymond Wieser
Case Western Reserve University
{yxf451,xxy530,rxw497}@case.edu

David Meakin
SunPower Corporation
david.meakin@sunpowercorp.com

Avishai Shaton
SolarEdge Technologies
avishai.shaton@solaredge.com

Jean-Nicolas Jaubert
CSI Solar Co.Ltd.
Jn.jaubert@csisolar.com

Robert Flottemesch
Brookfield Renewable U.S.
robert.flottemesch@luminace.com

Michael Howell
C2 Energy Capital
mh@c2.energy

Jennifer Braid
Sandia National Labs
jlbraid@sandia.gov

Laura S.Bruckman, Roger H.French, Yinghui Wu
Case Western Reserve University
{lsh41,rxf131,yxw1650}@case.edu

## ABSTRACT

The integration of the global Photovoltaic (PV) market with real time data-loggers has enabled large scale PV data analytical pipelines for power forecasting and long-term reliability assessment of PV fleets. Nevertheless, the performance of PV data analysis heavily depends on the quality of PV timeseries data. This paper proposes a novel Spatio-Temporal Denoising Graph Autoencoder (STD-GAE) framework to impute missing PV Power Data. STD-GAE exploits temporal correlation, spatial coherence, and value dependencies from domain knowledge to recover missing data. It is empowered by two modules. (1) To cope with sparse yet various scenarios of missing data, STD-GAE incorporates a domain-knowledge aware data augmentation module that creates plausible variations of missing data patterns. This generalizes STD-GAE to robust imputation over different seasons and environment. (2) STD-GAE nontrivially integrates spatiotemporal graph convolution layers (to recover local missing data by observed "neighboring" PV plants) and denoising autoencoder (to recover corrupted data from augmented counterpart) to improve the accuracy of imputation accuracy at PV fleet level. We have evaluated our proposed model on two real-world PV datasets. Experimental results show that STD-GAE can achieve a gain of 43.14% in imputation accuracy and remains less sensitive to missing rate, different seasons, and missing scenarios, compared with state-of-the-art data imputation methods such as MIDA and LRTC-TNN.

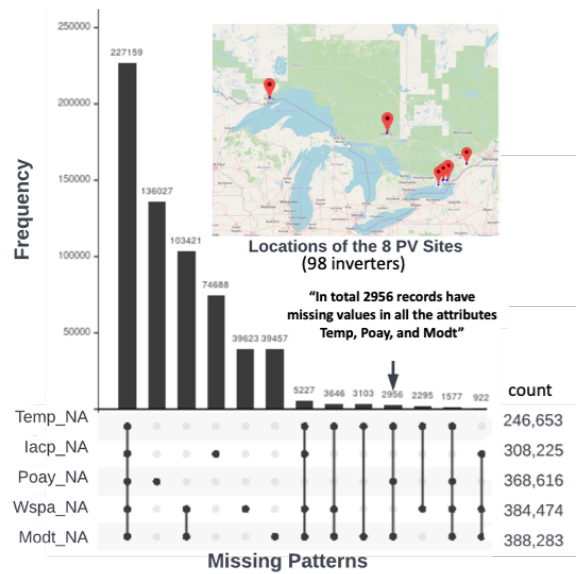

Figure 1: Illustration of most frequent (top 13 out of a total 27) missing data patterns and their frequency in a PV dataset of 98 PV inverters. Count: the number of records (*e.g.,* 246,653) having at least one (*e.g.,* temp) attribute value missing.

## 1 INTRODUCTION

Photovoltaics have become a dominant force in the energy sector over the past 20 years. Total installed solar capacity has now surpassed 773 GW and has grown by a factor of 500 since 2000 [8]. The exponential growth of the PV market has pushed the demand for power forecasting and performance evaluation for a huge population of PV power plants. Novel approaches seek to leverage sites which have spatiotemporal coherence that can be utilized for improving model accuracy [15]. However, real-time sensor measurements are prone to disruptions caused by measurement error, unexpected shutdowns of meters, equipment or component faults, power outrage, communication failures, etc [6]. These disruptions lead to single missing point and large blocks of missing data. The existence of missing data may impact the performance of downstream timeseries modeling such as the long-term photovoltaic degradation rate estimation [31].

Imputing missing timeseries data has been extensively studied. Notable examples include interpolation approaches [4, 21, 27], statistical learning [37, 39], or imputation with physical models [40]. Conventional PV data imputation is designed to impute the data for an individual PV plant or inverter, which often assume a certain missing data distribution (*e.g.,* missing at random), or require extra physical information to recovery missing data [40]. Deep models such as Graph Neural Networks (GNNs) have shown promising results in predicting timeseries data, such as traffic forecasting [10, 41–43]. While data imputation can be considered as a predictive task, existing models often assume high-quality, complete input data, a luxury that one does not have for real-world PV data.

There are several major challenges in PV data imputation:

- *Multiple missing scenarios*: The missingness of PV data may be characterized by different missing patterns, which are often hard to capture and impute with a single model;
- *Lack of high-quality input*: It is near impossible to obtain complete input as well as sufficient training examples, especially when there are multiple missing patterns;
- *Seasonality*: Even with the full observation of the timeseries data, the distribution of PV data vary due to seasonal variance, geospatial location, and transient weather phenomena.

For example, in our present dataset collected from 98 PV inverters (8 PV sites) that spans over three year (2014-2017), we found in total 27 different missing data patterns (determined by whether the value of a specific attribute is missing), as illustrated in Fig. 1. Among the top 13 most frequent patterns, the most frequent one indicates a "worst case" that all attribute values are missing. Additionally, 2.074% of the records have at least power value ("lacp") missing.

This example illustrates the need for an effective PV data imputation framework that can perform *unsupervised* and accurate imputation, bearing *noisy and incomplete* input, and with the presence of *multiple* missing patterns.

**STD-GAE Framework**. In response, we propose the *Spatio-Temporal Denoising Graph Autoencoder* (**STD-GAE**), a novel framework empowered by spatiotemporal GNNs to impute PV power data. Unlike conventional data imputation approaches, STD-GAE is optimized for PV data imputation with the following characteristics.

(1) STD-GAE is empowered by a spatiotemporal graph autoencoder to accurately learn PV network representation for missing PV data imputation. Our intuition is that PV fleet can be modeled as a spatiotemporally correlated inverter network. The measurement from one inverter can often help imputing its "similar" counterpart, which are captured by spatiotemporal correlations. To this end, STD-GAE adopts spatial, temporal, and inverter-level features in spatiotemporal graph convolution layers to learn PV fleet representations for imputation.

(2) To cope with sparse example, STD-GAE exploits a domain knowledge-aware data augmentation module. The module (a) leverages a suite of "plug-able" basic imputation methods to augment the input with imputed values, and (b) exploits a set of guard conditions from domain knowledge and physical models to validate the augmented input. This "cold-starts" STD-GAE learning with reasonable auxiliary data from even sparse PV observations.

(3) To generalize the imputation for PV input with multiple missing scenarios, STD-GAE takes a strategy of denoising autoencoders, whose goal is to learn accurate representation when part of input is missing, with an enhanced data corruption module, which allows configurable corruption with different missing types (*e.g.,* missing at random, block missing). By "enforcing" STD-GAE to reconstruct the corrupted yet augmented PV input, the imputation is able to achieve good performance for different missing scenarios.

Using real-world data collected from 98 PV inverters in Canada and a public PV dataset, we verified that STD-GAE can obtain a gain of 35.52% (resp. 15.09%) on average in MAE (resp. RMSE), compared to the state-of-the-art data imputation methods. The performance remains robust (not sensitive) even when the training of STD-GAE is constrained to a certain fraction of observations in a year.

**Related Work**. We summarize related work as follows.

*PV data imputation*. Conventional data-driven PV imputation usually adopts interpolation, statistical models or physical models. Notable examples include K-nearest Neighbor (KNN) and Linear Interpolation (LI), which interpolate missing data points by aggregating their neighboring ones[4, 21, 27]; and Multiple Imputation by Chained Equations (MICE) [37, 39], which uses statistical model and assumes missing at random pattern (MAR). These method only focus on imputing a single PV inverter or module, and often lead to biases if multiple missing patterns co-exists. Physical models have been proposed to use fully observed correlated attributes to recovery missing data in the target PV timeseries attribute [22, 40]. One of the major disadvantages is that if predictors are also missing, the model cannot sufficiently recover missing data. In addition, physical models rely on material and physical parameters that are highly variable and not well documented.

*Inverse Probability Weighting (IPW)*. IPW has also been proposed to deal with missing data. The basic idea is to construct a missingness model to estimate the probability that an individual is a complete case, and use it to adjust the biases induced by removing incomplete cases [34]. IPW and its evaluation is task-specific, usually designed for a downstream task *e.g.,* adult adiposity [24, 34, 35]. A possible downstream task of our imputation efforts is PV Performance Loss Rate (PLR) Prediction, where one can couple IPW with an imputation model and fine-tune it with unbiased datasets. However, we cannot directly compare IPW with our proposed imputation method, due to that IPW's design goal is quite different from directly minimizing imputation error.

*Spatiotemporal Graph Neural Networks (ST-GNNs)*. Recent advancement of Deep Learning, especially Convolutional Neural Networks (CNNs) leads to the rediscovery of Graph Neural Networks (GNNs). CNNs normally operate on structured data and images [45]. CNNs have been successfully used to impute missing in structured tabular data [18]. However, the PV network in question is an irregular graph where nodes (inverters) have different number of neighbors and feature distributions. Besides, the complexity of graphs impose challenges on existing CNNs based algorithms [38]. GNNs, a class of deep learning methods designed to perform learning tasks on graph data, often better capture the topological information within PV networks. STGNNs extends GNNs to model spatiotemporal networks with *e.g.,* recurrent graph convolutions [3, 36] or attention aggregated layers [42, 44]. The former captures spatiotemporal coherence by filtering inputs and hidden states passed to a recurrent unit using graph convolutions, while the latter learns latent dynamic temporal or spatial dependency through convolutions or attention mechanisms. Compared to STGNNs [41], we specify an expressive sandwiched

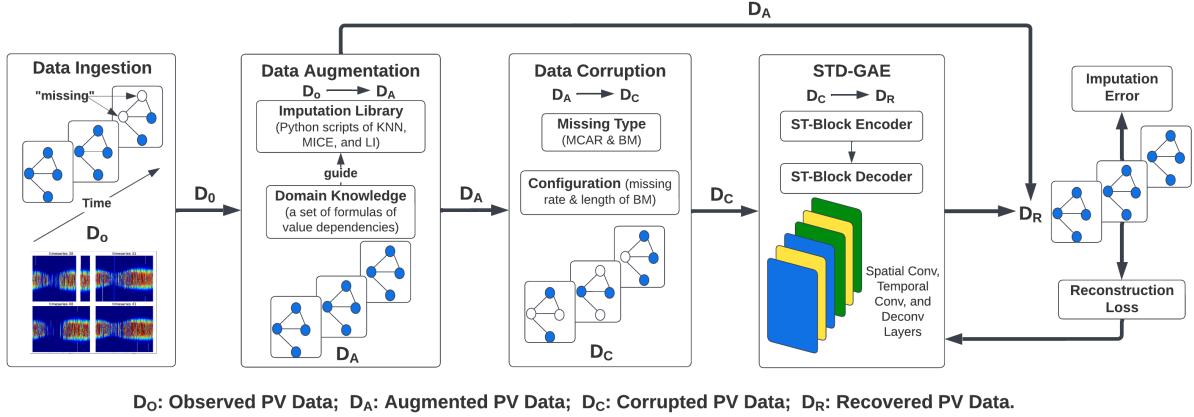

Figure 2: Overview of STD-GAE Imputation Framework.

$D_O$: Observed PV Data; $D_A$: Augmented PV Data; $D_C$: Corrupted PV Data; $D_R$: Recovered PV Data.

Spatio-temporal block, exploit data augmentation to improve the input quality, and support configurable data corruption to cope with diversified missing scenarios. To the best of our knowledge, this is the first work that integrates denoising autoencoders and spatiotemporal graph autoencoders for PV data imputation.

**Organization**. The remainder of this paper is structured as follows. Section 2 presents a brief introduction to Graph Neural Networks and Autoencoders and formulates the imputation problem. Section 3 introduces the proposed modular imputation method and provide details about each component. Section 4 represents and analyzes experimental results. Section 5 proposes application deployment of our model. Finally, the paper is concluded in Section 6.

## 2 PROBLEM STATEMENT

### 2.1 Graph Neural Networks and Autoencoders

**Graph Autoencoders**. Graph Autoencoders (GAEs) are unsupervised learning frameworks, consisting of a graph encoder and a graph decoder. The graph encoder learns network embeddings by mapping nodes into a latent representation, while the graph decoder learns to reconstruct the data from the encoded counterparts. Graph Convolution can be used to encode the nodes to produce a low dimensional network embeddings [19]. It aggregates signals from neighboring nodes to learn embeddings for each node:

$$\bar{X} = \sigma(\tilde{D}^{-\frac{1}{2}} \tilde{A} \tilde{D}^{-\frac{1}{2}} X W) \quad (1)$$

where $X$ is the node attribute matrix, $\tilde{A}$ is the adjacency matrix with self loop, $W$ is a trainable weight matrix, $\tilde{D}$ is the degree matrix with $D_{i,i} = \sum_j W_{i,j}$, and $\sigma$ denotes the activation function.

Previous studies in GAEs have shown their outstanding performance in learning from corrupted data, which is the natural extension of the problem of missing data imputation [28]. STD-GAE integrates GAEs to learn the distribution of node and topological features to handle complex missing scenarios.

**Denoising Autoencoders**. Denoising Autoencoder (DAE) is a variant of Autoendoer (AE), with a goal to recover data from corrupted input. AE tends to overfit for data imputation, since it minimizes the reconstruction loss between the input and the reconstructed counterpart that may lead to only learn an identity function. Instead, DAE introduces noise (corruption) to the input. Input data can be corrupted by noises added to the input vector in a stochastic manner [14]. The model is then trained to minimize the reconstruction loss between the recovered data and the uncorrupted counterpart.

### 2.2 PV Network Representation

**PV Network**. We represent the spatiotemporal PV data as an undirected graph $G = (V, E, X_t)$, where (1) each node in $V$ represents a PV inverter; and (2) $X_t$ denotes a node attribute tensor $\in \mathbb{R}^{T \times n \times d}$. Here T is the length of timeseries, $n$ is the number of nodes (which is 98 in our study), and $d$ is the number of input channel. Since the locations of PV inverters are fixed, the graph structure is static with time-invariant nodes and edges. However, $X_t$ is time-varying: each node i carries a timeseries $x_i \in \mathbb{R}^{T \times d}$ recording attributes such as temperature, wind speed, irradiance and power output.

*Modeling Edges*. We represent edges by edge index as a tensor $E_{index} \in \mathbb{R}^{2 \times m}$ and edge weight as a tensor $E_W \in \mathbb{R}^m$. Here $m$ is the number of edges. Both edge index and edge weight can be derived as follows:

$$W_{i,j} = \begin{cases} \exp\left(-\frac{d_{ij}^2}{\sigma^2}\right), i \neq j \text{ and } \exp\left(-\frac{d_{ij}^2}{\sigma^2}\right) \geq \epsilon \\ 0 \text{ otherwise} \end{cases} \quad (2)$$

where $d_{ij}$ is the Euclidean distance between the node pair $(i, j)$. $\sigma$ is the standard deviation of the distances. The network sparsity will be decided by $\epsilon$. When $\epsilon = 0$, we will have all nodes connected with each other. As $\epsilon$ increases, the sparser the network will be.

**PV data imputation**. Given a PV network $G$ with observed PV timeseries data $\{X_1, \ldots X_T\}$ as input, our goal is to impute missing power data with high accuracy. We characterize the task as a machine learning problem. Given a set of observed PV dataset $D_o$, our goal is to learn a graph autoencoder that minimizes the reconstruction loss between (a) a corrupted counterpart of an augmented version of $D_o$, and (b) the recovered data $D_R$ (see objective function Eq. 3). Here we corrupt and augment $D_o$ on purpose to avoid learning exactly the low-quality input, as remarked earlier.

We next introduce STD-GAE framework.

# 3 STD-GAE FRAMEWORK

We start with an overview of STD-GAE framework, and then present the details of its major modules: data augmentation, data corruption, spatio-temporal convolution and deconvolution.

## 3.1 Framework Overview

The STD-GAE framework, as illustrated in Fig. 2, consists of the following four major components.

**Data Ingestion**. STD-GAE first collects and transforms the raw input data from metering infrastructure of PV inverters to attributed data $D_O$. The PV data is stored in HBase supported by CRADLE, an HPC cluster at CWRU (see Section 5).

**PV Data Augmentation**. It then performs a data augmentation module to transform $D_O$ to augmented PV data $D_A$ from $D_O$ via a light-weighted imputation process. Hence, $D_A$ includes original $D_O$ and its augmented values. We remark that this stage is not to directly solve data imputation, but to provide relatively "more complete" data to improve the quality of the learned STD-GAE model. Specifically, (a) the imputation is cold-started by invoking a suite of basic imputation algorithms (as *e.g.,* user-defined functions) from a built-in Imputation library; (b) it then exploits domain knowledge, which is encoded as a set of value dependencies and rules, to validate and refine the imputed values.

**PV Data Corruption**. A data corruption module is performed to inject missing data to $D_A$, by declaring specified missing patterns and configuration. The corrupted data $D_C$ and the augmented data $D_A$ serve as the input to the training of STD-GAE model.

**Model Training**. In the model training phase, STD-GAE learns the denoising graph autoencoder model (simply referred to as STD-GAE) by minimizing reconstruction loss, i.e., the the mean squared error between the reconstructed data $D_R$ (given the corrupted input $D_C$) and $D_A$. Let the set of hyperparameters be $\Theta$, the value of recovered data for inverter $i$ at time $t$ as $D_R(t, i)$, the optimal parameter setting be $\theta$, training loss be $L(D_A, D_R)$, and number of observations be $N = T \times n$, the learning objective is to minimize the following loss function:

$$\theta = \arg\min_{\theta \in \Theta} \frac{1}{N} \sum_{i=1}^{T} \sum_{t=1}^{n} (D_R(t, i) - D_A(t, i))^2 \quad (3)$$

Note that the reconstruction loss is different from imputation error since the latter is only computed on the missing part of test data. Instead of focusing only on the reconstruction loss of missing data in the training stage, our trained model learn spatiotemporal correlations from the whole training data to better recover missing in the out-of-sample data. We train, validate, test our model in a sliding window fashion with both window size and step size set to be 288. Since each interval is five minutes, a window of size 288 amounts to a sequence of measurements with a length of one day.

## 3.2 Data Augmentation and Corruption

**Data Augmentation with Domain Knowledge**. Data augmentation module aims to improve the imputation accuracy by generating more labels, cold start STD-GAE training, and mitigate the negative impact of low-quality data. STD-GAE benefits from properly enriched datasets. While the data augmentation does not solve the PV data imputation alone, it is to properly fill in the missing data detected in data ingestion step to obtain a "refined" observed dataset $D_A$ for training purpose. STD-GAE supports a "plug-able" Imputation Library that contains python scripts of a set of primitive imputation operators such as linear interpolation (LI), KNN, and MICE. It helps to generate more labels for STD-GAE training. As such, the augmented data together with observed real-world data form the ground truth in the training.

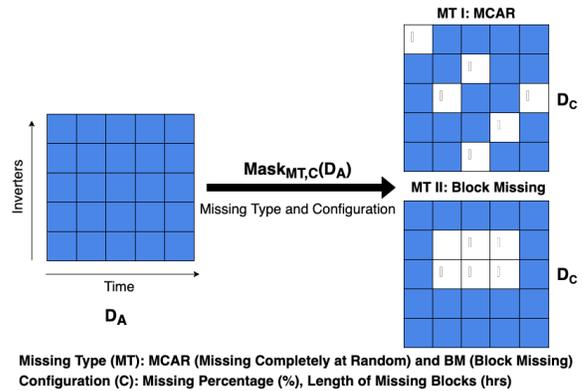

**Missing Type (MT)**: MCAR (Missing Completely at Random) and BM (Block Missing)
**Configuration (C)**: Missing Percentage (%), Length of Missing Blocks (hrs)

**Figure 3: PV Data Corruption.**

STD-GAE leverages domain knowledge from PV scientists and engineers to prevent augmented values from deviating largely from reality. Domain knowledge is encoded as a set of formulas of value dependencies among PV timeseries attributes. For example, in PV industry, a simple Predicted Power model following the irradiance and temperature scaling approach is widely acknowledged [11].

$$P = \frac{G_{POA}}{1000} \frac{P_{norm}}{1 + \gamma_T (T_{module} - 25)} \quad (4)$$

where $P$ is the estimated power, $P_{norm}$ is the power of the PV module at standard testing condition, $G_{POA}$ is the irradiance incident on the plane of the module or array ($W/m^2$), $\gamma_T$ is the temperature coefficient of PV modules, and $T_{module}$ is the module temperature (°C). While this heuristic rule does not fully reflect all deterministic factors of $T_{module}$ (*e.g.,* it may ignore the impact of low light, clipping loss, shading, and module degradation), STD-GAE leverages the *value dependencies* to validate impossible values. For example, the above equation is used to set a reasonable value ranges (domains) for the missing power timeseries data to be imputed, along with other value constraints for *e.g.,* nameplate power of PV modules: $P_{nameplate} \geq P_{norm} \geq 80\% P_{nameplate}$; since a typical solar panel's performance warranty will guarantee 80% at 25 years. In our tests, K-nearest Neighbor achieved satisfiable performance that introduces few violations of the rule constraints, and are chosen as a major imputation operator by default. Fig. 4 illustrates PV power output values of 30 randomly sampled inverters before and after the validated data augmentation.

The data augmentation module can be easily extended to similar scenarios by updating the current framework with corresponding domain knowledge. For example, constraints can be posed on wind power, hydropower, and residential/commercial utilities consumption from the energy industry.

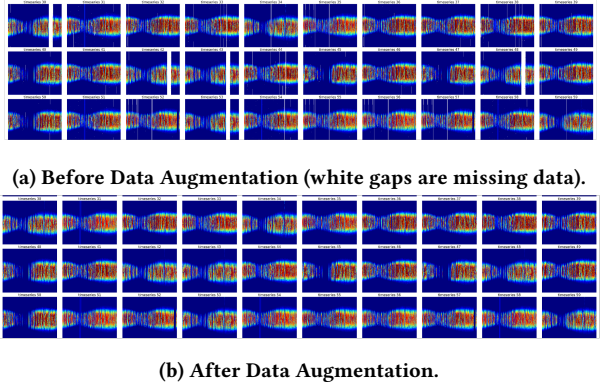

(a) Before Data Augmentation (white gaps are missing data).

(b) After Data Augmentation.

Figure 4: Heatmaps of Daily Power Output of Randomly Sampled 30 PV Inverters.

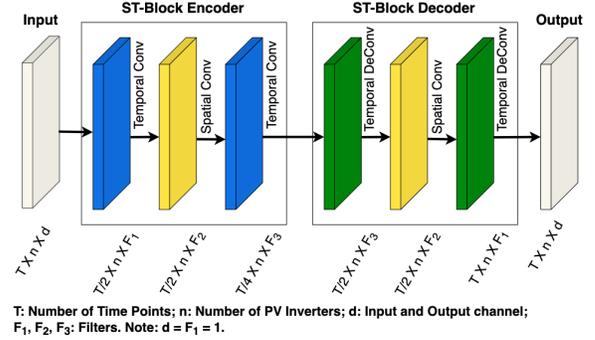

T: Number of Time Points; n: Number of PV Inverters; d: Input and Output channel; $F_1$, $F_2$, $F_3$: Filters. Note: $d = F_1 = 1$.

Figure 5: Structure of the Spatial Layers and Temporal Layers in the Proposed STD-GAE.

**Data Corruption.** In the Data Corruption module, we generate missing masks to simulate the distribution of real-world missing patterns of PV data. Missing mask is characterized by a configuration tuple $(MT, C)$, where MT denotes a set of desired missing types (*e.g.,* MCAR or BM), and C denotes a configuration parameters: missing rate or length of block missing.

Fig. 3 shows an example of corruption phase using the above two missing patterns, where the white (resp. blue) blocks denote missing (resp. augmented) values. (1) For MCAR, missing data masks are generated using a uniform distribution between 0 and 1 such that the threshold corresponds to the selected configuration of missing percentage. All the missing values are randomly scattered which is typically caused by short-term power or communication failure. Thus, they are independent of each other. (2) For BM, missing data masks are created by injecting a fixed length to daily timeseries of each PV inverter. The fixed length is chosen according to the selected configuration of length of block missing. BM is more commonly caused by a longtime malfunction of the sensors.

According to the selected missing mask $Mask_{MT,C}$, we inject missing data points to $D_A$. If $Mask_{MT,C}(t, i) = 1$, then $D_C(t, i)$ is observed, while $D_C(t, i)$ is missing if $Mask_{MT,C}(t, i) = 0$. We represent corrupted data $D_C$ as $D_C = Mask_{MT,C}(D_A) = D_A \odot Mask_{MT,C}$, where $\odot$ is the element-wise product operator.

### 3.3 Spatial-Temporal Blocks

We next detail our design of spatial layers and temporal layers in the ST-Blocks of STD-GAE.

**Spatiotemporal (ST)-Block.** Each ST-Block consists of one spatial layer sandwiched between two temporal layers. Both decoder and encoder are composed of a ST-Block. As illustrated in Fig. 5, the ST-Block encoder extracts the PV power characteristics by mapping input data into lower dimensional node embeddings, while the ST-Block decoder recovers original dimension of the reduced data.

*Temporal Layers.* To capture the temporal correlations, i.e., correlations in the timeseries of each PV inverter, we use gated 1D convolution in both encoder and decoder. A 1D filter $\Omega \in \mathbb{R}^K$ is used to perform convolutions on daily timeseries of each PV inverter by aggregating neighboring values in timeseries. $\Omega * x$ performs convolution on $x$ such as its length is reduced to a half with $K = 4$, stride = 2, and padding = 1. Then a gated linear unit (GLU) is used as activation defined as:

$$GLU = (\Omega * x) \odot \sigma(\Omega * x) \quad (5)$$

where $\sigma$ is a sigmoid function and $\odot$ is element wise product operator. After the temporal convolutional layers in encoder, deconvolutional layers in decoder restore length of timeseries to original size by performing reverse convolutions. Temporal convolutional and deconvolutional layers are designed in a way such that input and output of the STD-GAE have the same size.

*Spatial Layers.* A spatial convolutional layer performs convolution on a graph by aggregating data points from neighboring nodes in both encoder and decoder. We adopt the chebyshev spectral graph convolutional operator (ChebConv) [9] as our spatial convolutional layer. ChebConv is a spectral-based method. Let $L = I_n - D^{-\frac{1}{2}}AD^{-\frac{1}{2}}$ be the normalized graph Laplacian matrix. Since L is real symmetric definite, it can factored as $L = U\Lambda U^T$, where $U \in \mathbb{R}^{n \times n}$ is the matrix of eigenvectors ordered by eigenvalues and $\Lambda$ is the diagonal matrix of eigenvalues with $\Lambda_{ii} = \lambda_i$. The spectral convolution of a filter kernel $\mathbf{g}_\theta$ with signal X is defined as:

$$X *_G \mathbf{g}_\theta = U\mathbf{g}_\theta U^T X \quad (6)$$

where $U^T X$ is a graph Fourier transform to signal X. Since the eigen-decomposition requires $O(n^3)$ computational complexity, ChebConv reduces the complexity to $O(m)$. ChebConv approximates $\mathbf{g}_\theta$ by Chebyshev polynomials of $\Lambda$, i.e., $\mathbf{g}_\theta = \sum_{i=0}^{k} \theta_i T_i(\tilde{\Lambda})$, where $\tilde{\lambda} = \frac{2\Lambda}{\lambda_{max}} - I_n$. Since $T_i(\tilde{L}) = UT_i(\tilde{\lambda})U^T$, it can be shown that Equation 3 can be estimated by:

$$X *_G \mathbf{g}_\theta = \sum_{i=1}^{K} \theta_i T_i(\tilde{L})X \quad (7)$$

where $\tilde{L} = \frac{2L}{\lambda_{max}} - I_n$ and $k$ is the size of the kernel for spatial convolution. The value $K$ shows signals passed to nodes up to $k$ hops away. In this study, we choose $k = 3$ to achieve a balance between cost and accuracy.

# 4 EXPERIMENTAL STUDY
## 4.1 Experimental Setup
**Dataset**. We verify our model on two real-world PV datasets. (1) A private powerplant PV dataset contains normalized inverter alternating current power of 98 PV inverters from 8 PV sites in Canada. It covers 360 days ranging from 09/01/2015 to 08/25/2016. (2) The second dataset[1] is a public residential PV dataset with 35 inverters from 3 PV sites in Phoenix Arizona. For a fair comparison, we calibrated the latter by taking a consecutive fraction over the same historical period, consistently from 09/01/2015 to 08/25/2016.

We split both datasets into training, validation, and testing as follows: day 1-240 for training, day 241-300 for validation, and day 301-360 for testing. For a fair comparison of all methods, we set the training and validation "ground truth" as a subset $D_{G'} \subseteq D_A$, and the testing "ground truth" a subset $D_{G''} \subseteq D_O$. Our dataset contains sample interval of 5 minutes, amounting to 103,680 data points for each PV inverter. Around 1.22% entries of the ingested private PV dataset are natually missing while there are no missing data in the ingested public PV dataset. Most of missing values in private dataset are either random missing (only one value missing) or block missing (continuous chunks of missing). In data corruption, missing values are injected based on configured selected missing type and fraction. For example, 50% MCAR means 50% values are missing and distributed randomly.

**Evaluation Metrics**. We use two classes of evaluation metrics.

*Imputation accuracy*. The imputation accuracy is evaluated by Mean Absolute Error (MAE) and Rooted Mean Squared Error (RMSE), and domain knowledge based metrics. For each tested missing scenario, a data corruption mask selected according to the selected missing data type (MCAR or BM) and configuration (missing rate or length of missing interval). We denote test data as $D_T$, the observed part of it as $D_{T_o} \subseteq D_T$, and the augmented test data as $D_{T'}$. The data corruption (mask) is only applied to a subset of $D_{T'}$ such that all the elements have counterparts in $D_{T_o}$. Denote $D_{T'_c} \subseteq D_{T'}$ as the corrupted data, imputed values of $D_{T'_c}$ as $P$, and the ground truth of $D_{T'_c}$ as $\tilde{P}$. Each metric can be defined as follows:

$$MAE = \frac{1}{N}\sum_{i=1}^{N}|P_i - \tilde{P}_i|; \quad RMSE = \sqrt{\frac{1}{N}\sum_{i=1}^{N}(P_i - \tilde{P}_i)^2} \quad (8)$$

where $N = card(D_{T'_c})$, $P_i \in P$, and $\tilde{P}_i \in \tilde{P}$.

*Domain-specific Metrics*. We also consider *percentage of outliers* and *seasonality*, two established domain-specific metrics for quantifying the quality of imputed PV data [13, 16, 23, 29]. Specifically, we evaluate whether the imputation algorithms can "recover" the same level of outlier percentage and seasonality in the original dataset. (1) The outliers refer to the data points that are greater than ± 1.5 times the interquartile range. We adopt tsoutliers R package to detect outliers [25]. (2) It is natural to investigate whether the imputation methods are able to maintain seasonality, which is a common feature of PV timeseries. We use the STL function to decompose the each dataset into a seasonal component, a remainder component, and a trend using locally weighted nonparametric regression, and compare the seasonal component among datasets from various imputations. The STL functions are available in the base R stats package. Both outlier detection and seasonality characterization functions are available to our PVplr R package [1, 2].

[1]https://datahub.duramat.org/dataset/phoenix

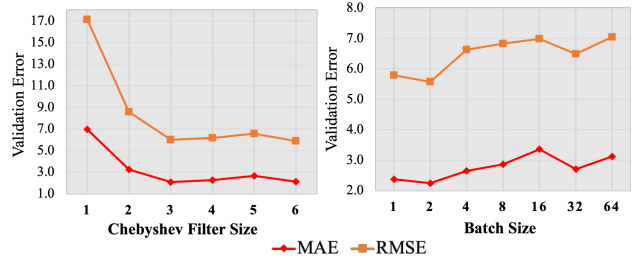

**Figure 6: The Impact of Chebyshev Filter Size and Batch Size on Validation Error.**

**Imputation Methods**. We compare STD-GAE with 6 baselines.

(1) *Linear Interpolation (LI)* [5]: is a timeseries imputation method that fits a simple linear model using two values before and after the missing data block. Each missing data point will then be estimated using the linear model.

(2) *Mean Imputation (Mean)* [12]: is a common approach that uses the column-wise mean to fill the missing data.

(3) *KNN* [26]: imputes data by finding and averaging $K$ nearest neighbors to fill in the missing value.

(4) *MICE* [32]: makes multiple imputations using chained equations. It imputes missing values of a dataset by a divide and conquer approach. Once the focus is placed on one variable, MICE uses all the other variables in the dataset to predict missing part of that variable. The prediction is based on a regression model, with the form of the model depending on the nature of the focus variable.

(5) *LRTC-TNN* [7]: Low-Rank Tensor Completion with Truncated Nuclear Norm minimization (LRTC-TNN) is one of the state-of-the-art spatiotemporal imputation technique designed initially for traffic data imputation. It formulates the data imputation problem in a low-rank tensor completion (LRTC) framework and defines a novel truncated nuclear norm (TNN) on spatiotemporal data in form of *location* × *day* × *time*. In particular, it introduces an universal rate parameter to control the degree to allow better characterize the hidden patterns in spatio-temporal data.

(6) *MIDA*: [14]: is a denoising autoencoder approach that uses fully connected layers. Unlike the denoising graph autoencoders in our proposed STD-GAE, the encoder in MIDA maps the input data to a higher dimension while decoder maps back to original input. The model is trained on randomly corrupted inputs.

**Configuration**. STD-GAE is trained by Adam Optimizer [20]. We have used grid search to find the optimal setting of hyperparameters in partitions of their discrete grid space[30]. We implement a grid search with the following setting of parameters: ChebShev filter size $\in \{1, 2, 3, 4, 5, 6\}$, $\epsilon \in \{0, 0.25, 0.5, 0.75, 1.0\}$, batch size $\in \{1, 2, 4, 8, 16, 32, 64\}$, and epoch $\in \{25, 50, 100\}$. We set learning rate to be 0.001 adjusted with a decay rate of 0.02. In Fig. 6, we use 6-hours BM as an example to illustrate how validation error change when we vary either batch size or chebyshev filter size. Based on our validation results, we choose ChebyShev filter size to be 3 and batch size to be 2. The network sparsity parameter $\epsilon$ is set to be 1 (please also refer to Fig. 11). The number of epochs is 50. We have also studied the impact of model depth of STD-GAE on imputation accuracy. When we increase the number of ST-Block to be 4 from 2 (see Fig. 5 for our current design of layers), RMSE and MAE of STD-GAE increase 30.73% and 75.62% separately. This result verifies that adding more ST-Blocks (layers) degrades the model performance, and justifies our design of STD-GAE.

All experiments are implemented with Pytorch Geometric Temporal [33] and conducted on Intel Xeon(R) CPU E5-2630 v4 @ 2.20GHz, 64 GB Memeory, 6 CPU cores, and 12GB NVIDIA GeForce RTX 2080 GPU. Our source code, public PV dataset, and a full version of the paper are made available [2].

## 4.2 Experiment Results

We next introduce our findings.

**Exp-1: Comparisons of Imputation Accuracy**. We investigate two missing data types as mentioned in the Section 3.2, i.e., missing completely at random (MCAR) and block missing (BM). We conduct controlled experiments by varying the missing rate of MCAR from 10% to 60% and length of missing intervals of BM from 2 hours to 12 hours. In total, we test 12 different missing scenarios.

*Accuracy (private data)*. Fig. 7 reports MAE and RMSE of proposed method and baselines over the private dataset. We have the following observations. (1) STD-GAE achieves the best imputation performance in most scenarios. Second best LRTC-TNN achieves similar performance with STD-GAE only in the test RMSE of BM, but worse than STD-GAE in other cases. (2) All imputations perform better in MCAR than BM. Compared to MCAR, the increase of imputation error is the highest for LI and the lowest two are STD-GAE and LRTC-TNN. This finding indicates that STD-GAE can leverage spatial correlations of PV inverters to recover the large chunks of missing data when neighboring timestamps are not available. (3) For the same missing data type, as the severity (missing percentage or length of block missing) increases, STD-GAE maintains a comparable gain since it is trained by minimizing the reconstruction loss of whole training data to learn a better overall distribution of PV power data. Note that results of Mean Imputation are not shown here due to its errors being out of scale, i.e., delivering the worst performance in every missing scenario.

*Accuracy (public data)*. We also tested STD-GAE on the public residential PV data that is distributed vast differently from its counterpart in Fig. 7. We can observe that STD-GAE achieves either best or top performance in all twelve different missing scenarios that we have tested, as shown in Fig.8. STD-GAE remains less sensitive to the increase of missing rate and length of missing block which is consistent to results in Fig. 7.

*Visual analysis*. We randomly select two PV inverters from the private data and visualized the results for two missing scenarios: 40% MCAR (left hand side) and 6-hours BM (right hand side) in Fig. 9, with validated ground truth in one-day period. The left hand side figure depicts a window with 40% random missing data. STD-GAE can recover most of the missing observations and capture the majority of sharp fluctuations. The right hand side of Fig. 9 illustrates a counterpart with 6-hours block missing. For this window, STD-GAE accurately recovers 6-hours missing block by utilizing the spatial coherence from neighboring inverters.

**Exp-2: Comparison with Domain Knowledge-Based Metrics**. To investigate whether the imputations could recover the outliers mainly induced by cloud shading, the difference between the recovered data using seven different imputations and the "ground truth" - pseudo imputed test data from Data Augmentation for each individual inverter, are illustrated in Fig. 10a and 10b. Except for Mean and MICE, all methods are able to recover outlier percentage in MCAR. While all algorithms tend to generalize the results, introducing outliers to several inverters in locations that were far less often shaded by clouds, only STD-GAE and LRTC-TNN perform well in BM, agreeing with the results in Fig. 7.

Fig. 10c and 10d verify the difference in seasonality feature between imputed data and test data. LI works well for MCAR but poorly for BM. For the two missing types, STD-GAE and LRTC-TNN outperform the rest. For 12-hours BM, LRTC-TNN outperforms STD-GAE at maintaining seasonality for places with large variance in power output between seasons, e.g., snow coverage reduced significant power loss in winter as compared to other seasons. On the other hand, STD-GAE utilizes seasonality features from neighboring inverters which are similar to each other, and performs best in most cases. In Fig. 10d, all data points are below 0 and none is able to recover exactly the same seasonality. This is due to a "worst case" where training and testing data are from different seasons (with quite different distributions). We are obtaining larger-scale PV datasets with length of more than 3 years to overcome this issue.

**Exp-3: Case Studies.** We conduct case analyses to validate the robustness of our proposed STD-GAE on different seasons and evaluate the impact of distant PV inverters/sites on imputation error. We also conduct ablation analyses to evaluate the impact of imputation library and domain knowledge in augmentation module.

*Seasonality*. We verify the robustness of our trained STD-GAE imputation model on different seasons by selecting four out-of-sample monthly PV data, including October 2016, January 2017, April 2017, and July 2017 as representatives of four seasons. We choose the missing type of 40% MCAR as the example for verification. The improvements in test RMSE compared to the closest baseline method are 27.79%, 23.77%, 19.27%, and 30.52% separately for these four months. Although the STD-GAE is not trained on a whole year's PV data, we still achieve relatively stable improvement in imputation accuracy for representative months from all seasons.

*Network topology*. We also investigate the impact of network sparsity to the imputation accuracy at both fleet (graph)-level and inverter (node)-level. As $\epsilon$ increases, the graph has fewer edges (see Eq. 2) since edges connecting distant PV sites and inverters are filtered. As the dataset report that inverters at the same site share the same longitude and latitude, when $\epsilon = 1$, we obtain the sparsest graph as 8 "cliques", where only inverters from the same site are pairwisely connected. We choose 6-hours BM as the example for comparison. As shown in Figs. 11 and 12, STD-GAE obtains the smallest imputation error on both fleet-level and inverter-level when $\epsilon = 1$, where there are no edges between inverters if they are from different sites. This justifies our selection of edge models. While it indicates that distant PV sites and inverters tend to reduce the fleet-level and inverter-level imputation accuracy, it is possible that remote site "contributes" to the imputation of a local one due to similar environmental and weather conditions. We leave this for further study over more datasets. On the other hand, the "optimal" $\epsilon$ may change upon the change of physical networks (*e.g.,* moving or removing inverters). STD-GAE can be readily updated accordingly.

---
[2]https://anonymous.4open.science/r/STD-GAE-B8FD

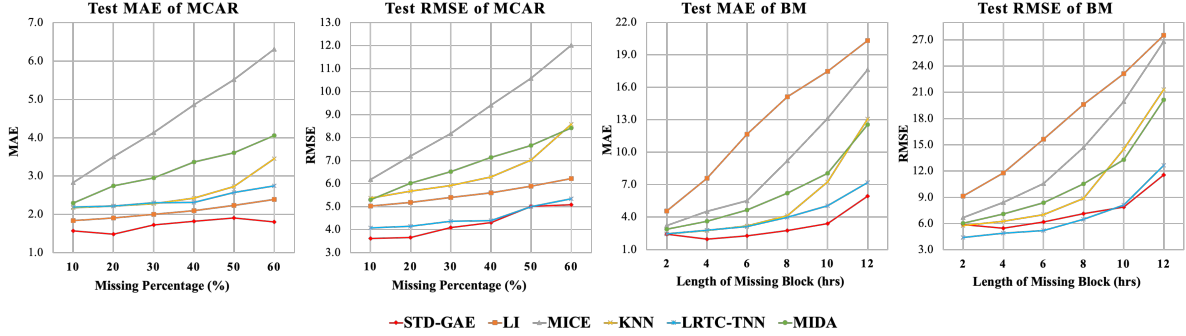

Figure 7: Imputation Errors and Impact of Missing Scenarios and Severity (results of Mean Imputation are out of scale).

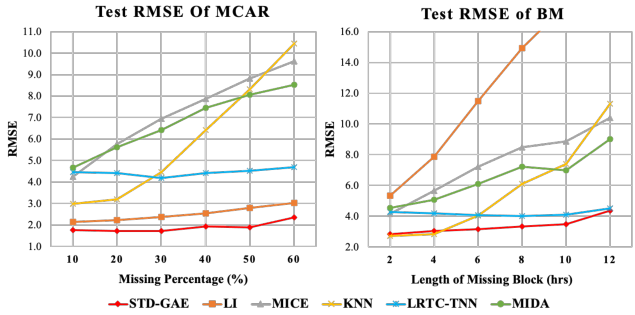

Figure 8: Imputation Errors and Impact of Missing Scenarios and Severity (public PV dataset).

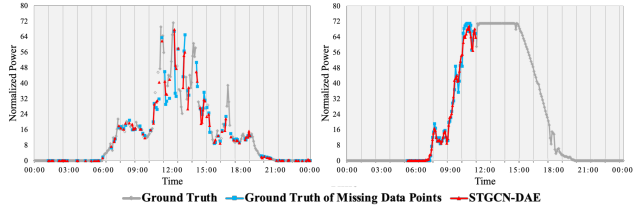

Figure 9: Imputation Results of the Proposed STD-GAE (left: 40% MCAR, right: 6-hours BM)

*Ablation analysis.* To verify how the augmentation module may improve data imputation, we conduct two ablation analyses to study the impacts of imputation library and domain knowledge. (1) We first select two missing scenarios, 30% MCAR and 6-hours BM, to evaluate domain knowledge. We found that imputation with domain knowledge validation reduces MAE and RMSE by 6.70% and 6.11% separately on average for the counterpart without domain knowledge. (2) We next choose 50% MCAR and 12-hours BM to study the impacts of imputation library. The augmentation with base imputation reduces the test MAE and RMSE by 3.06% and 3.47% respectively on average, in these two cases. These results verified that incorporating augmentation module with domain knowledge can improve imputation accuracy.

## 5 FRAMEWORK DEPLOYMENT

The deployment of our proaposed STD-GAE imputation model involves two steps as shown in Fig. 13.

**PV Data Management**. The PV data management consists of three steps: data acquisition, data preprocessing, and storage, similar to the previous work [17]. First, data comes from various commercial PV power plants. We collect these datasets through web APIs, secure shell FTP, or receiving them as CSV files over the cloud as encrypted zip files. We create cron jobs to run particular file parsers for datasets with different formats. Second, in data preprocessing, once data arrives, the first task is anonymization. We anonymize the proprietary information and save the anonymized data in Hadoop cluster. In the data processing step, timeseries from HDFS are read and passed through validation, tidying, and uniform structuring. Numerical values are checked for missing percentage or anomalies and assigned the quality score. Finally, only timeseries with high-quality scores are ingested into HBase. Data in HBase are stored in a cell such that the value in a cell is uniquely identified by row, column qualifier, column family, and timestamp. To address the large overhead caused by millions of rows, the database is designed such that each cell contains a large amount of data compared to its unique identifier and keeps one month of data as a string in cell.

**Model Deployment**. STD-GAE will be deployed to improve PV datasets for commerical plants, which will be integrated into the data preprocessing. Acquisition of real-time data at regular intervals is crucial for imputation. The deployment of a production pipeline has to been integrated with the data management and inference modules. Data from HBase will be processed with Data Augmentation and Augmentation Modules, two python packages developed internally to provide high-quality data for STD-GAE training and evaluation. STD-GAE will be trained on high performance clusters.

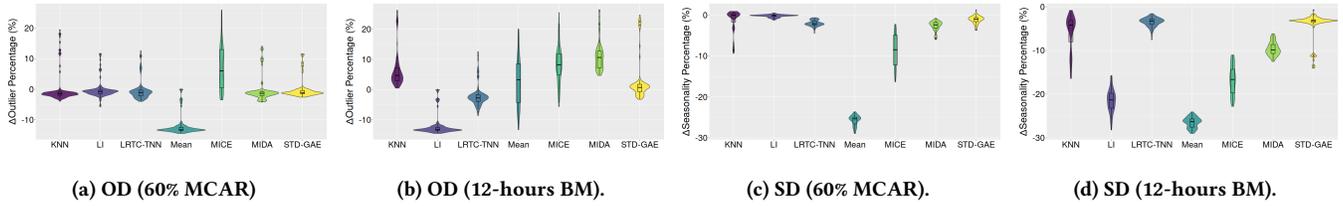

(a) OD (60% MCAR)    (b) OD (12-hours BM).    (c) SD (60% MCAR).    (d) SD (12-hours BM).

Figure 10: Comparison of Recovery of Domain Knowledge-Based Metrics (OD: outlier difference and SD: seasonality difference) Using Different Imputation Methods for Two Missing Types.

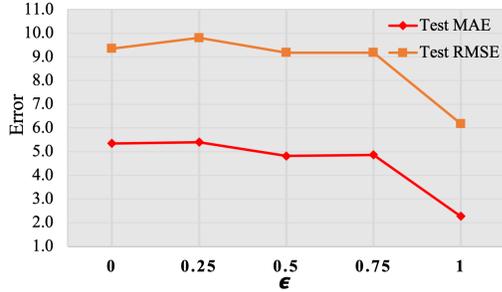

Figure 11: Impact of Network Sparsity on Fleet-level Accuracy.

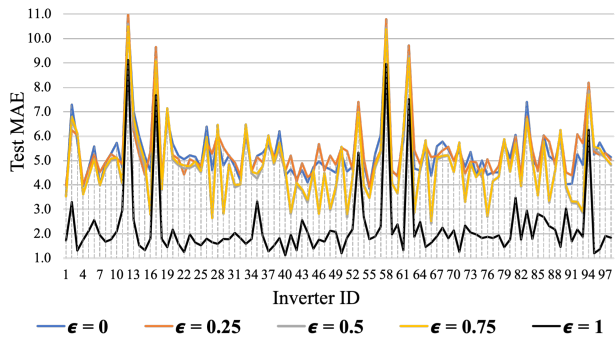

Figure 12: Impact of Network Sparsity on Inverter-level Accuracy.

The imputed ("recovered") data $D_R$ will be forwarded to our PV commercial collaborators and research PV farms.

Our proposed framework is generally applicable to other domains that involves spatiotemporal data imputation, e.g., missing data imputation in wind power generation, utility electricity, carbon dioxide emission, nitrogen cycling in agriculture, traffic network, etc. Domain knowledge from corresponding fields can easily guide the data augmentation module and trains STD-GAE accordingly.

## 6 CONCLUSION AND FUTURE WORK

In this study, we have proposed Spatio-Temporal Denoising Graph Autoencoder (STD-GAE) that combines temporal correlations within-series and spatial correlations from neighboring PV inverters to recover missing data. Since STD-GAE is trained to minimize the reconstruction loss of the corrupted input, we can better learn spatiotemporal correlations and data distribution to recover missing data. We compare STD-GAE with existing methods in a real-world dataset from 98 PV inverters in Canada and a public PV dataset. STD-GAE obtains an improvement of 35.52% and 15.09% on average in the test MAE and test RMSE compared to the state-of-the-art missing data imputation methods like MIDA and LRTC-TNN. As the missing rate of MCAR or the missing block size of BM increases, the performance gap between the STD-GAE and baseline imputations become larger. Our tests also verify that STD-GAE retains data properties such as percentage of outliers and seasonality.

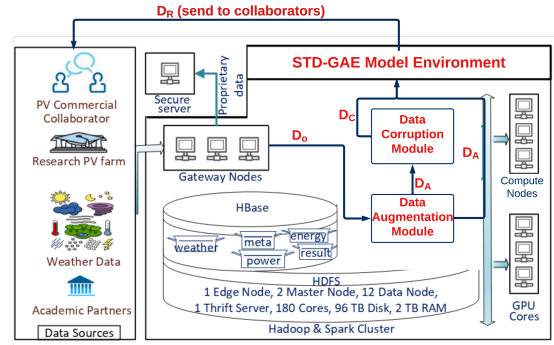

Figure 13: Proposed STD-GAE Imputation Workflow in CRADLE. A cron job periodically executes a program to ingest PV data from different sources into Hadoop cluster. Data Augmentation and Data Corruption modules are the internal packages developed to interact with HBase and STD-GAE imputation model. Adapted from [15].

A future topic is to train STD-GAE on our newly ingested PV data in HBase that covers a span of at least three consecutive years. Another topic is to incorporate STD-GAE and IPW to handle missing data in our downstream task PV Performance Loss Rate (PLR) Prediction and enable multiple channels for attributes temperature and irradiance in the PV production model, to further improve the performance of PV data imputation and degradation analysis.

## ACKNOWLEDGMENTS

This material is based upon work supported by the U.S. Department of Energy's Office of Energy Efficiency and Renewable Energy (EERE) under Solar Energy Technologies Office (SETO) Agreement Number DE-EE0009353. We thank Didier Thevenard from Canadian Solar Inc. for providing data.